\documentclass[letterpaper, 10 pt, conference]{ieeeconf}  

\usepackage{amsmath}
\usepackage{amssymb}
\usepackage{graphics} 
\usepackage{epsfig} 
\usepackage{mathptmx} 
\usepackage{times} 
\usepackage[font={small,it}]{caption}
\usepackage{hyperref}
\usepackage{wrapfig}
\usepackage{authblk}
\usepackage[caption=false,margin=0pt]{subfig}
\usepackage{algorithm}
\usepackage{algpseudocode}
\usepackage{svg}
\IEEEoverridecommandlockouts                              

\title{Deep Object-Centric Representations for Generalizable Robot Learning\\
}

\author{
  Coline Devin$^1$, Pieter Abbeel$^{1,2}$, Trevor Darrell$^1$, Sergey Levine$^1$%
  \thanks{$^1$UC Berkeley, Department of Electrical Engineering and Computer Science}%
  \thanks{$^2$ OpenAI}%
  \thanks{\tt coline,pabbeel,trevor,svlevine@eecs.berkeley.edu}%
}

\begin{document}
\maketitle

\thispagestyle{empty}
\pagestyle{empty}

\begin{abstract}
Robotic manipulation in complex open-world scenarios requires both reliable physical manipulation skills and effective and generalizable perception. In this paper, we propose a method where general purpose pretrained visual models serve as an object-centric prior for the perception system of a learned policy. We devise an object-level attentional mechanism that can be used to determine relevant objects from a few trajectories or demonstrations, and then immediately incorporate those objects into a learned policy. A task-independent meta-attention locates possible objects in the scene, and a task-specific attention identifies which objects are predictive of the trajectories. The scope of the task-specific attention is easily adjusted by showing demonstrations with distractor objects or with diverse relevant objects. Our results indicate that this approach exhibits good generalization across object instances using very few samples, and can be used to learn a variety of manipulation tasks using reinforcement learning.
\end{abstract}

\section{Introduction}
Recent years have seen impressive improvements in the performance of computer vision systems, brought about by larger datasets~\cite{russakovsky2015imagenet}, improvements in computational capacity and GPU computing, and the widespread adoption of deep convolutional neural network models~\cite{he2015deep}. However, the gains in computer vision on standard benchmark problems such as ImageNet classification or object detection~\cite{lin2014coco} do not necessarily translate directly into improved capability in \emph{robotic} perception, and enabling a robot to perform complex tasks in unstructured real-world environments using visual perception remains a challenging open problem. 

\begin{figure*}
\centering
\includegraphics[width=\textwidth]{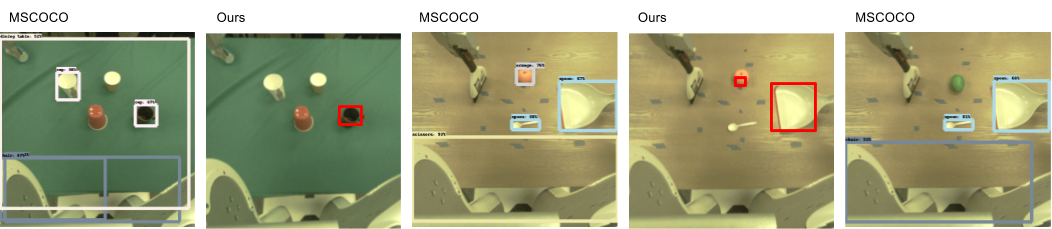}
\caption{Right: Faster RCNN trained on MSCOCO does not differentiate between cups and mugs, and gives higher probability to the cup, making it difficult to train a policy that needs to locate the mug. With our method, the attention can learn to prioritize mugs over cups. Left: The dustpan is labeled as a spoon and thus can be distracted by other spoon-like objects. As limes are not in the MSCOCO dataset, the object detector does not label them.}
\label{fig:mscoco}
\end{figure*}

Part of this challenge for robot perception lies in the fact that the effectiveness of modern computer vision systems hinges in large part on the training data that is available. If the objects for a task happen to fall neatly into the labels of dataset, then using a trained object detector for perception makes sense. However, as shown in Figure~\ref{fig:mscoco}, objects outside the label space may be labeled incorrectly or not at all, and objects that the robot should distinguish may be labeled as being the same category. If the robot's environment looks too different from the detector's training data, the performance may suffer. 
These difficulties leave us with several unenviable alternatives: we can attempt to collect a large enough dataset for each task that we want the robot to do, painstakingly labeling our object of interest in a large number of images, or we can attempt to use the pretrained vision system directly, suffering from possible domain shift and a lack of flexibility. In both cases, we also have limited recourse when it comes to correcting failures: while we can add additional labeled data if the robot fails on a particular object, a single new labeled data-point is unlikely to correct the mistake.

\begin{figure}[t]
  \centering
  \includegraphics[width=0.5\textwidth]{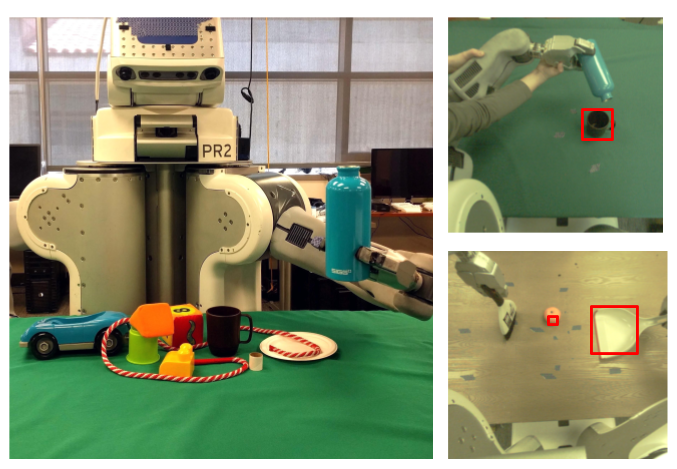}
  \caption{Deep object-centric representations learn to attend to task-relevant objects from just a few trajectories. The representation is robust to clutter and generalizes to new object instances.}
  \label{fig:teaser}
\end{figure}

An alternative view of robotic vision has emerged in recent years with advances in deep reinforcement learning~\cite{jaderberg2016reinforcement,mnih2016asynchronous}, end-to-end learning from demonstrations~\cite{daftry2016learning,xu2016end}, and self-supervised learning~\cite{pinto2016curious, pinto2016supersizing,levine2016learning}. These methods bypass the standard computer vision representation of class labels and bounding boxes and directly train task-specific models that predict actions or task success from raw visual observations. While these methods can overcome the challenges associated with large-scale semantic labeling and dataset bias by training directly on the task that the robot aims to solve, their ability to generalize is critically dependent on the distribution of training data. For example, if a robot must learn to pour liquid from a bottle into a cup, it can achieve instance-level proficiency with a moderate number of samples~\cite{finn2016guided}, but it must train on a huge number of bottles and cups in order to generalize at the category level. Switching from the standard vision framework to end-to-end training therefore allows us to bypass the need for costly human-provided semantic labels, but sacrifices the generalization that we can get from large computer vision datasets.

In this work, we seek to develop a robotic vision framework that operates on sets of objects rather than raw pixels, and leverages prior datasets to learn a generic object concept model.
Our principal insight is that, if the robot will be using learning (e.g., reinforcement learning or learning from demonstration) to perform the final task that is set out before it, it does not require precise labels or segmentation. It simply needs to be able to consistently localize visual cues in the observed image that correlate with the objects that are necessary for it to perform the task. However, to learn generalizable policies, the visual cues should be semantic in nature such that a policy trained on one object instance can function on a similar object when desired. 

We therefore take a two-stage approach to robotic vision: in the first stage, we construct a   object-centric attentional prior based on an region proposal network. This stage requires a large amount of data, but does not require any task-specific data, and can therefore use a standard existing computer vision dataset. The second stage narrowly focuses this general-purpose attention by using a very small number of example trajectories, which can be provided by a human or collected automatically during the reinforcement learning process. This teaches the system to attend to task relevant objects, while still benefiting from the generalizable representations present in the general-purpose attention. Furthermore, because the second stage is trained with only a handful of example trajectories, it makes it easy for the user to correct mistakes or control the class of objects that the system generalizes to, simply by providing additional demonstrations. For example, if the user needs a policy specific to a particular type of cup, they can simply provide demonstrations with other cups present in the scene, illustrating that they should be ignored. If the user prefers broader category-level generalization, for example to cause a robot generalize across all types of fruits, they might provide demonstrations that show interactions with fruits of different types. In all cases, the total number of provided trajectories remains very small (less than 15).

The main contribution of our work is a perception framework that facilitates generalization over objects and environments while requiring minimal data or supervision per task. 
 Our method incorporates general-purpose object-centric priors in the form of an object attention trained on a large, generic computer vision dataset, and combines it with an extremely efficient task-specific attention mechanism that can either be learned from a very small number of demonstrations, or even specified directly by a human operator. We show that this framework can be combined with reinforcement learning to enable a real-world robotic system to learn vision-based manipulation skills. Our experiments demonstrate that our approach achieves superior generalization to an end-to-end trained approach, through the incorporation of prior visual knowledge via the general-purpose attention. We  illustrate how the user can control the degree of generalization by including or excluding other objects in the demonstrations. Our source code is available online in a stand-alone ROS package: \url{https://github.com/cdevin/objectattention}. A video of results is available here: \url{https://sites.google.com/berkeley.edu/object-representations}.

\section{Related Work}
Vision for robotics is often approached differently from general computer vision as it involves interacting directly with the environment~\cite{bohg2017interactive}. Robot perception is often concerned  with object detection and localization at the instance level, such as through 3D representations~\cite{schmidt2015depth}~\cite{hinterstoisser2011multimodal}, keypoints~\cite{lowe2004distinctive}, or deep neural networks~\cite{held2016robust}. Modern computer vision approaches to detection tend to use deep neural networks trained on large datasets labeled at the category level~\cite{ren2015faster,he2017mask}. Category-level reasoning is appealing because it can generalize across object instances, but is limited by the labels included in a dataset. Our approach allows the level of generalizing to be learned and modified from the objects seen during in the given trajectories. 

Our method combines prior knowledge about ``objectness'' from pretrained visual models with an attentional mechanism for learning to detect specific task relevant objects. A number of previous works have sought to combine general objectness priors with more specific object detection in the context of robotics and other visual tasks. Ekvall et al. used region proposals and SIFT features for quickly detecting objects in the context of SLAM~\cite{ekvall2007object}. Prior work used an active search approach where the camera could zoom in certain parts of the receptive field to search at higher resolutions~\cite{kawanishi2002quick}.
In manipulation, SIFT features have also been used  for 3D pose estimation and object localization, using object-specific training data gathered individually for the task~\cite{collet2009object,tang2012textured}. Similarly to these prior works, our method constrains the observations using an object-centric prior. However, we do not require object level supervision for each task, instead using visual features from a pretrained visual model to index into the proposals from the object-centric prior. This approach drastically reduces the engineering burden for each new task, picking out task-relevant objects from a few demonstrations, and provides good generalization over object appearance, lighting, and scale, as demonstrated in our experiments.

\begin{figure*}[h]
  \centering
  \includesvg[width=\textwidth]{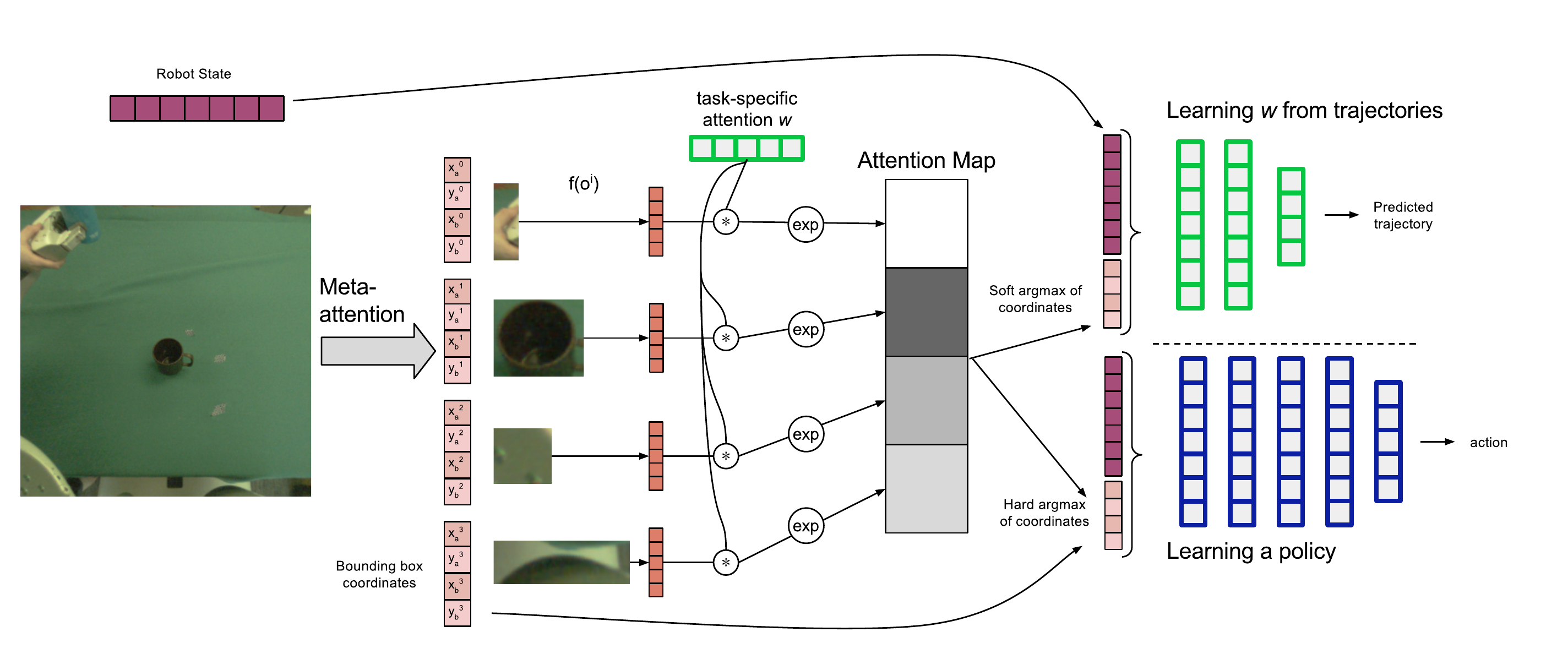}
  \caption{Method Overview. The parameters outlined in bright green
  are optimized during pretraining, while those outlined in dark blue are optimized during policy learning. The attention is trained to predict the movement seen in the provided demonstrations or trajectories. The ``attention map" illustrates a soft attention where vectors $f(o^i)$ close to $w$ are given high probability (black) and ones far away are have low probability (white). The distribution  is regularized to have low entropy, and the weighted sum of bounding box coordinates is fed to the next layers of the bright green network. The policy (in blue) is trained with  $w$ held fixed, and the arg-max bounding box is used instead of the weighted average. Note: this diagram illustrates only a single attention vector $w$; more attention vectors can be added as needed.}
  \label{fig:method}
\end{figure*}

An alternative to building perception systems for task-specific objects is to learn the entire perception system end-to-end together with the control policy. A number of recent works have explored such end-to-end approaches in the context of skill learning, either for direct policy search~\cite{levine2016end,pinto2016curious,pinto2016supersizing}, unsupervised learning of representations for control~\cite{ghadirzadeh2017deep,finn2016deep}, or learning predictive models~\cite{agrawal2016learning,finn2017foresight}. A major challenge with such methods is that their ability to generalize to varied scenes and objects depends entirely on the variety of objects and scenes seen during policy training. Some methods have sought to address this by collecting large amounts of data with many objects~\cite{pinto2016supersizing,levine2016learning}. In this work, we instead study how we can incorporate prior knowledge about objects from pretrained visual models, while still being able to train rich neural network control policies. In this way, we can obtain good generalization to appearance, lighting, and other nuisance variables, while only training the final policy on a single object instance and in a single scene.

\section{Deep Object Centric Representations}
The goal of our method is to provide a simple and efficient process for quickly acquiring useful visual representations in the context of policy learning. Specifically, we aim to compress an image into a vector $\nu$ of object descriptors and select task-relevant objects from it. We impose an object-centric structure on our representation, which itself is learned from prior visual data in the form of standard computer vision image datasets. We define a 2-level hierarchy of attention over scenes for policy learning. The high level, which we
call the \textbf{meta-attention}, is shared for all tasks. The meta-attention is intended to identify possible objects
in the scene regardless of the task. The lower level, which we call \textbf{task-specific attention}, is learned per-task and identifies
which of the possible objects is relevant to the task being performed.

\begin{algorithm}
\caption{Robot Learning with Object Centric Representations}
\label{alg}
\begin{algorithmic}[1]
  \State Train meta-attention on an object detection dataset (shared for all tasks).
  \State Train a convolutional network $f$ for image classification (shared for all tasks). 
  \For {each task} 
  \State Collect demonstrations.
  \State Learn task-specific attention $W$ as described in Section~\ref{sec:lfdattention} from the trajectories.
    \State Train control policy using reinforcement learning, with the robot's configuration and $\nu$ as the observation inputs. $W$ is fixed during this step.
  \EndFor
\end{algorithmic}
\label{algo}
\end{algorithm}

\subsection{Meta-Attention}
\label{sec:lfdattention}
The meta-attention is a function that takes an image and returns a set of object hypotheses $\{o^i : i \in [0, N)\}$. Each object hypothesis consists of a semantic component (given by $f(o^i)$) and a position component (given by $g(o^i)$). The semantic component describes the identity of the object with a feature vector, while the position component describes where the object is located within the image. Importantly, this meta-attention is reused over all tasks without retraining. Its proposals are task-independent and it constitutes our object-centric perception prior.

Although a number of meta-attention mechanisms are possible, we use a region proposal method to provide a set of possible objects.
The objects $o^0, ..., o^N$ are the proposed crops and the position component $g(o^i)$ are the bounding box coordinates of the proposal.
We define the semantic component $f$ to be a mean-pool over the region proposal crop of the convolutional features pretrained on ImageNet classification~\cite{russakovsky2015imagenet}. Because $f$ is convolutional, it does not require a particular input size, so can feed in each crop as is. As the convolutional layers were pretrained for classification with a diverse dataset, the vector will include information about the contents of the
crop (e.g., it may have ``mug-like'' features) that should be invariant to translation, rotation, illumination, and other variables.

\subsection{Task-Specific Attention}
\label{sec:learningattention}
Once the task-independent object hypotheses are obtained, we use a task-specific attention to choose objects to attend to for a given task.
In the tasks we examine, an object's relevance is determined by its identity: if the task is to pour into  a mug, the object that looks
like a mug is the relevant object. To select which objects to pay attention to, the model learns the \textit{task-specific} attention over the semantic features $f(o^i)$. 
In order to be able to quickly learn the task-specific attention from only a small number of trajectories, it is parametrized as a vector $w$ such that attention paid to $o^i$ is proportional to $e^{w^{\top}f(o^i)}$. To attend to several different objects, multiple attention vectors can be stacked into a matrix $W$.

Because the attention is linear with respect to the semantic features, the choice of these semantic features is crucial for the flexibility and generalization capability of the method. While we could choose the features to simply correspond to semantic class (e.g., using classes from a standard image classification dataset), this would limit the flexibility of the method to identifying only those object classes. If we choose overly general features, such as a histogram of oriented gradients or even raw pixels, the task-specific attention would be too limited in its ability to generalize in the presence of changes in appearance, lighting, viewpoint, or pose. To strike the right balance between flexibility and generalization, we use visual features obtained from the upper layers of a convolutional neural network trained for image classification. Such features have previously been shown to transfer effectively across visual perception tasks and provide a good general-purpose visual representation~\cite{simonyan2014two,sermanet2013overfeat}. An overview of our method is provided in Algorithm~\ref{algo}.

The learnable weights in the local attention are the values of $W$, which attend over the visual features of each crop.
$W$ should learn to identify what kinds of object to pay attention to for a given task, and the objects relevant to a task should be predictive of successful trajectories performing the task.
The aim of this section is to train $W$ on trajectories to attend to task-relevant objects. Once $W$ is learned, any reinforcement learning algorithm could be used to obtain a policy as a function of the attended objects.

Given trajectories of a task, the objects that are relevant to the task will be predictive of future robot configurations. Trajectories could come from a variety of sources; in this paper we use either kinesthetic demonstrations or directly make use of the trajectories obtained during reinforcement learning (in our case, with guided policy search).
We optimize for $W$ as part of a larger neural network shown at the top of Figure~\ref{fig:method} that aims to predict the next step in the trajectory: an action if available, or a change in position of the end-effector. 
The network for this is two hidden layers with 80 units each. In order to backpropagate through $W$, a soft attention is used. First, we use a Boltzmann distribution to obtain a probability $p(o^i|w_j)$ for each object proposal.
\[p(o^i|w_j) = \frac{e^{w_j^{\top} \frac{f(o^i)}{||f(o^i)||_2} }}{\sum_{i=0}^N e^{w_j^{\top}\frac{f(o^i)}{||f(o^i)||_2} }}\]
Then, the soft attention is calculated by taking a weighted sum of the object locations.
\[\nu_{j,\text{soft}} = \sum_{i=0}^N g(o^i)p(o^i|w_j).\]
To obtain the prediction, $\nu_{soft}$ is concatenated with robot joint state and end-effector state before being fed into the movement prediction network at the top of Figure~\ref{fig:method}. 
While $f(o^i)$ is normalized for each $o^i$, $W$ is not normalized to allow the optimization to control the peakiness of the attention. To encourage more discrete attention distributions, the attention is regularized to have low entropy.
\[\mathcal{L}_{\text{ent}}(w) = \sum_{j=0}^M \sum_{i=0}^N -p(o^i|w_j)\log p(o^i|w_j)\]
The network is optimized with the Adam optimizer~\cite{kingma2014adam}.

To better condition the optimization when the task-relevant objects are known, the task-specific attention can be  initialized by providing one example crop of the desired object(s) before finetuning on the demonstration data. 

\begin{figure*}
  \centering
  \includegraphics[width=\textwidth]{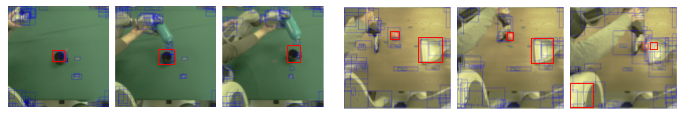}
  \caption{The region proposals (meta-attention) are drawn in blue and the task-specific attended regions are drawn in red. For the pouring task, the attention locks on to the mug as its position defines the trajectory. For the sweeping task, we use two attention vectors, one attends to the orange and one attends to the dustpan, which each have variable starting positions. }
  \label{fig:attention}
\end{figure*}
\section{Experiments}
We evaluate our proposed object-centric model on several real-world robotic manipulation tasks. The experiments are chosen to evaluate two metrics: the reliability of this representation for robotic learning, and how well it generalizes to visual changes in the environment. By attending over features trained on the diverse images found in ImageNet, we expect that policies learned with our visual representation will naturally generalize to visual changes. The hard attention over region proposals should provide robustness against distractor objects.
The aim of this evaluation is to demonstrate that the model enables both policy learning and generalization to new object instances and environments. Additionally, we show that the scope of generalization can be modified by showing different objects during demonstrations, which is particularly useful for correcting mistakes that the attention might make.

\subsection{Training details}
The meta-attention is provided by a region proposal network (RPN)~\cite{ren2015faster} trained on the MSCOCO dataset~\cite{lin2014coco}. For the semantic component of each object, we use conv5 of AlexNet~\cite{krizhevsky2012imagenet}, resulting in a 256-dimensional feature vector which is then normalized to have magnitude 1.  Videos of the results can be found at \url{https://sites.google.com/berkeley.edu/object-representations}.

The attention vector $w$ is learned by training a model on trajectory data as described in Section~\ref{sec:learningattention}. The attended regions learned for both tasks are shown in Figure~\ref{fig:attention}. To learn to perform the task, we use the guided policy search algorithm~\cite{levine2014learning}, which involves training local time-varying linear-Gaussian controllers for a number of different instances of the task, which in our case correspond to different positions of the objects, and then using supervised learning to learn a single global neural network policy that can perform the task for all of the different object positions. The neural network policy takes as input the joint angles, joint velocities, end-effector positions and velocities, as well as the output of the perception system $\nu$, which corresponds to the attended region's bounding box coordinates. The learned policies have 4 hidden layers with 80 hidden units each, and directly output the torques for the 7 joints of the PR2 robot used in our experiments. Note that our representation can be used with any reinforcement learning algorithm, including both deep reinforcement learning methods (such as the one used in our experiments) and trajectory-centric reinforcement learning algorithms such as PI2~\cite{TheodorouBS10} or REPS~\cite{PetersMA10}.

\label{sec:pouringtask}
\subsection{Generalizing across visual changes}

In this experiment, we evaluate the hypothesis that attending over high-level features trained on classification will generalizes across objects of the same class.
The goal of the this task is to position a bottle to pour into a mug. Success requires the ability to locate a mug from an image and the global policy is not given the mug location, and we use different mugs for training and test. We
compare against a task-specific approach from Levine et al. which learns the policy directly from raw pixels with a spatial softmax architecture\cite{levine2016end}. While optimizing perception directly for the task objective performs well on particular mug seen during training, our method can generalize to new mug instances and to cluttered environments. Although the method in Levine et al. pretrains the first convolutional layer on ImageNet, conv1 features are too low-level to provide semantic generalization.

\begin{figure}[t]
  \centering
      \includesvg[width=0.45\textwidth]{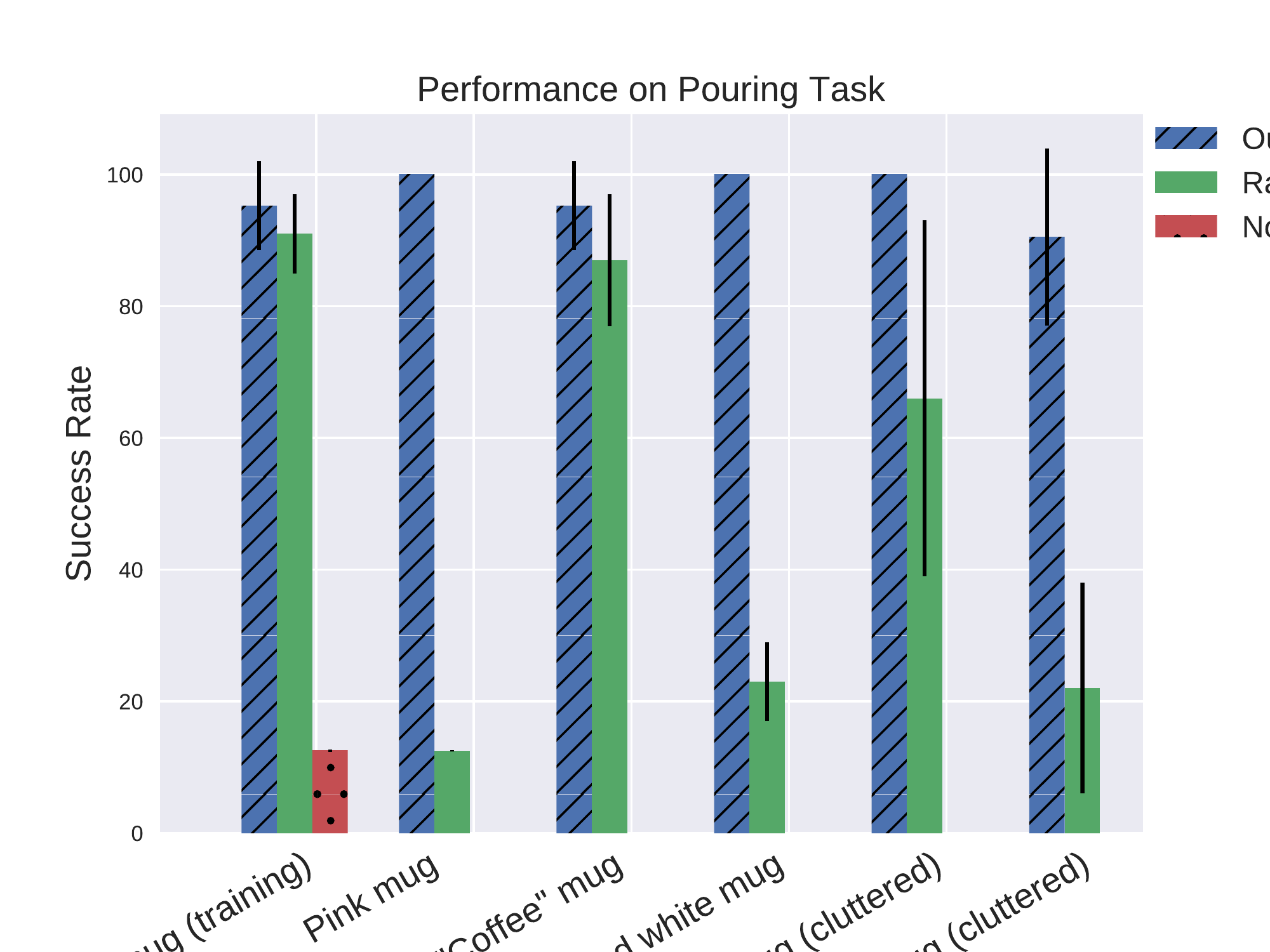}
  \caption{Results for the pouring task testing on different mugs not seen during training. Each group of bars shows performance on different unseen mugs, comparing our method with a policy trained from raw pixels. Our policy successfully generalizes to a much wider range of mugs than the raw pixel policy, which does not benefit from the object-centric attention prior. The ``no vision" baseline indicates the performance of always pouring at the average mug position.}
  \label{tab:pouring}
\end{figure}

\begin{figure}[t]
  \centering
  \includegraphics[height=0.20\textwidth]{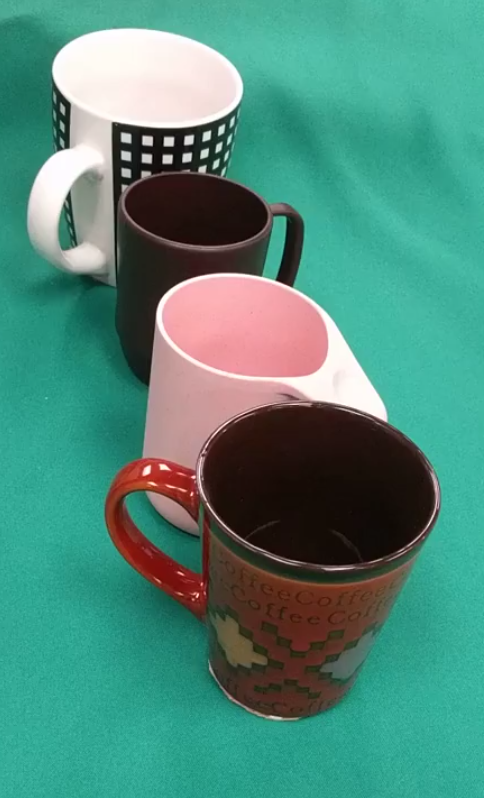}
  \includegraphics[height=0.20\textwidth]{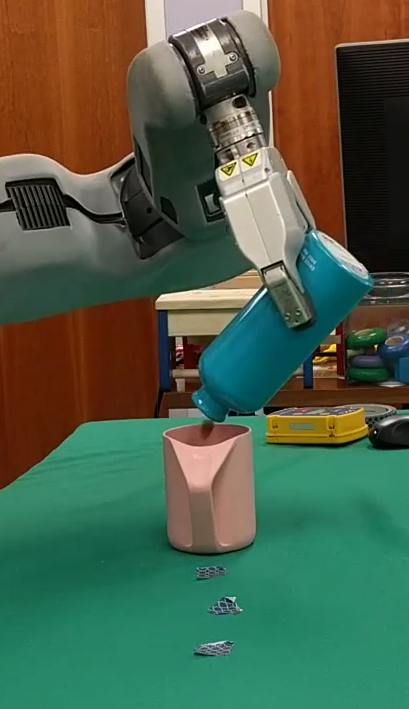}
  \includegraphics[height=0.20\textwidth]{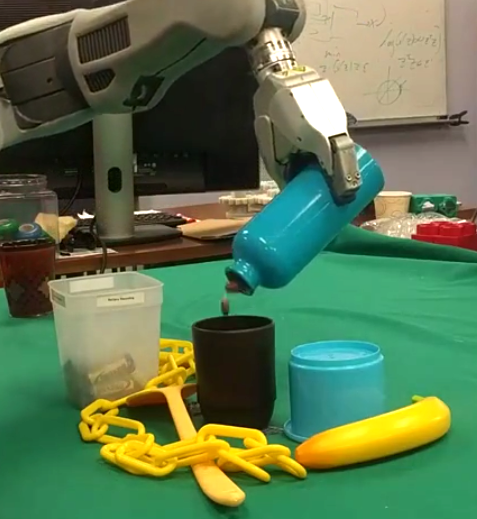}
  \caption{Left: Mugs used for evaluation. Note that only the brown mug was seen during training. Center: Successful pouring into the pink mug. Right: Pouring into the brown mug in a cluttered environment that was not seen during training.}
  \label{fig:mugs}
  \end{figure}

For evaluation, the policy is run with almonds in the bottle. A rollout is marked as successful if more almonds fall into the mug than are spilled, as seen in the included video. For evaluation, eight rollouts at different mug positions are performed for the uncluttered environments and three for the cluttered ones; results are in Figure~\ref{tab:pouring} and environment photos are in Figure~\ref{fig:mugs}. 

While the policy was only trained on a single brown mug in a plain environment, it successfully generalizes to other mugs of various colors. By using hard attention, the visual features are robust to clutter. Interestingly, when presented with all four mugs, the policy chose to pour into the pink mug rather than the brown mug the attention was trained with. The ``no vision" baseline is a policy trained without visual features; its behavior is to pour to the average of the different targets seen during training. The low performance of this baseline indicates that the task requires a high level of precision. We compare to the method described in~\cite{levine2016end}, where policies are learned directly from raw pixels and pretrained on a labeled data  for detecting the target object.

Our model is able to generalize to new mugs of different appearances because it uses deep classification features that were trained on a wide variety of objects including mugs. An approach that learns robot skills directly from pixels such as~\cite{levine2016learning} could not be expected to know that the brown mug and the pink mug are similar objects. We investigate this by training a policy from raw pixels with the architecture described in ~\cite{levine2016learning}. The convolutional layers are pretrained on detecting the training mug in 2000 labeled images. As shown in Table~\ref{tab:pouring}, this policy can perform the task on the training mug and on another brown mug, but completely ignores the other mugs. This indicates that a policy learned from raw pixel images can perform well on the training environments, the kinds of features it pays attention to have no incentive to be semantically meaningful and general. Our method of using features pretrained on image classification defines how a policy should generalize. 

\subsection{Learning to ignore distractors}
In the first experiment, generalizing across mugs was a desired outcome. However, it is easy to imagine that a task might require pouring specifically into the brown mug and ignoring all other mugs. Our method provides a simple way for the user to adjust which features the task-specific attention is sensitive to. In order to learn a task-specific attention that has a narrower scope, the user can simply add another mug -- in this case, a pink mug -- as a distractor during the demonstrations. As described in Section~\ref{sec:learningattention}, the vector $W$ is trained such that the attended box allows predicting the arm's movement in the demo trajectories. As the pink mug is not predictive of the trajectories, the gradient pushes the attention vector to lock on to brown mug specifically. We used 6 additional demonstrations to finetune the attention.

\begin{figure*}
  \centering
  \subfloat[Left: The soft attention from just training on the brown mug is shown in red. Right: The soft attention after finetuning on demonstrations where the pink mug is present. When initialized on only the brown mug, the attention is sensitive to "mug" features, and therefore can be distracted by the pink mug. After adding demonstrations of pouring into the brown mug with the pink mug in the background and finetuning, the attention has locked on to just the brown mug. The blue squares show the meta-attention.]{\label{subfig:mug}
  \includegraphics[width=0.754\textwidth]{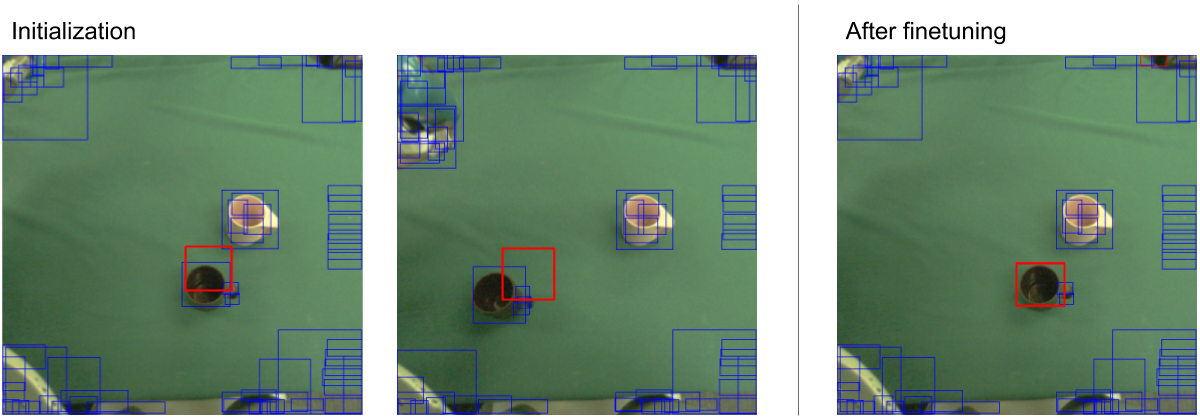}}\\
\subfloat[Top: The soft attention in red from just training on the orange. Bottom: The soft attention after finetuning on demonstrations of sweeping oranges, lemons, and limes. Although the attention was initially sensitive to "orange-specific" features, finetuning on other fruit made the attention generalize to lemons and limes.The blue squares show the meta-attention.]{\label{subfig:citrus}\includegraphics[width=0.9\textwidth]{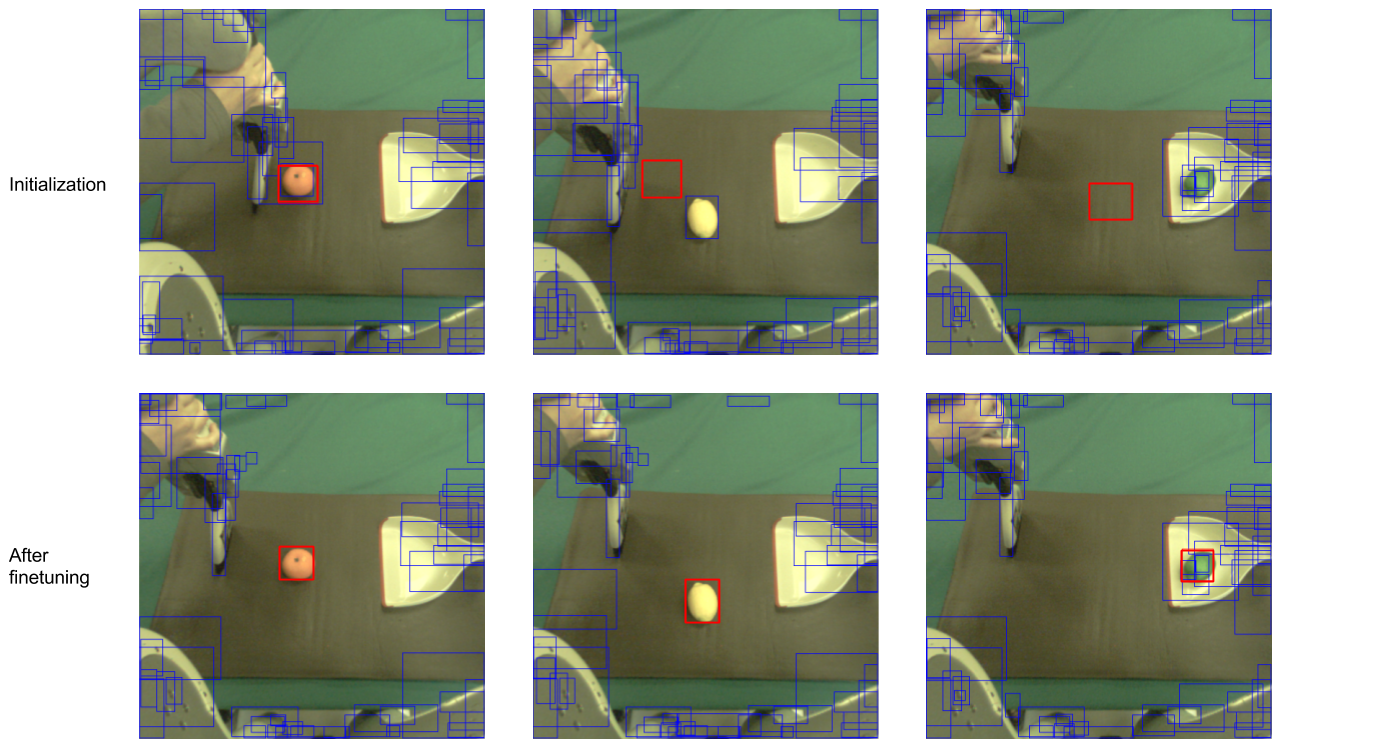}}
  \caption{}
\end{figure*}

At test time, the pouring policy consistently chooses the brown training mug over both the pink mug and the black-and-white mug. This indicates that including distractors in the demonstrations helps restrict the scope of attention to ignore these distractors. Figure~\ref{subfig:mug} shows how an attention initialized just on the brown mug is distracted by the distractor mug. After finetuning on the demonstrations, the attention is firmly on the brown mug. In experiments, the robot poured into the correct mug 100\% of the time with either the pink mug or the black and white mug present as distractors. In comparison, the attention trained solely on demonstrations without distractors preferred the pink mug over the brown mug and obtained 50\% success. This experiment shows that if a roboticist were to find that the attention vector is over-generalizing to distracting objects, it is easy for them to gather a couple more demonstrations to narrow down the attention. 

\subsection{Increasing the scope of generalization}
Since our method is not limited by the labels present in available datasets, the attention vectors can also be pushed to attend to a greater variety objects. For example, a vector that attends to oranges may not always generalize to other citrus fruit. However, if generalizing across citruses is desired, the user can easily correct this mistake by adding a couple demonstrations with limes and lemons and finetuning $W$.
In comparison, if a researcher relying on an off-the-shelf detector were to disagree with the detector's performance, modifying the model could require relabeling data the model was trained on or collecting and labeling new detection data. 

As shown in Figure~\ref{subfig:citrus}, the attention only attends to oranges when first initialized, but finetuning expands the scope of the attention to include the lime and lemon present in the demonstration data. The resulting sweeping policy is robust to distractors including an apple, apricot, and a purple cup, but is confused by the orange  and green cups. The round base of the citrus-colored cups perhaps appear to be idealized fruit.

\subsection{Learning to attend to multiple objects}
In this experiment we demonstrate that we can use multiple attention vectors to learn a policy that depend on the location of two objects. The robot learns to perform a sweeping task where the object to be swept (a plastic orange) and the dustpan each can start in different positions. Ten kinesthetic demonstrations were collected to learn a pair of attention vectors, initialized with a single crop of the objects. 
This policy successfully sweeps the orange into the dustpan for 77\% of the trials. The policy is robust to distractors and works even if the dustpan is held by a person. As shown in the video, this task is difficult because the robot's angle of approach needs to be a function of the relative positions of the orange and dustpan, and the orange changes in appearance as it rolls. A baseline policy which did not use images at all succeeded at 11\% of the test positions indicating that visual perception is necessary for this task.

\section{Discussion}
In this paper, we proposed a visual representation for robotic skill learning that makes use of object-centric priors from pretrained visual models to achieve robust perception for policies trained in just a single scene with a single object. Our approach uses region proposal networks as a meta-attention that picks out potential objects in the scene independently of the task, and then rapidly selects a task-specific representation via an attentional mechanism based on visual features, which can be trained from a few trajectories. Since the visual features used to index into the object proposals are themselves invariant to differences in lighting, appearance, viewpoint, and object instance, the resulting vision system can generalize effectively to new object instances with trivial additional training. The attention's scope is easily controlled able by the objects seen during demonstrations. Our results indicate that this provides for a robust, generalizable, and customizable visual representation for sensorimotor skills. This representation generalize across different mugs when trained on only one mug, but could also be instance-specific if shown a handful of additional trajectories. In the opposite case, we show that an attention that was narrower than desired could be broadened as needed. Finally, for tasks that require interacting with multiple objects we can learn multiple attention vectors that and sensitive to different objects.

While our method attains good results on two real-world manipulation tasks, it has a number of limitations. First, the visual representation that is provided to the policy is constrained to correspond to image-space object coordinates. Although this is sufficient for many manipulation tasks, some tasks, such as those that require fine-grained understanding of the configuration of articulated or deformable objects, might require a more detailed representation. Second, our current system requires pretraining on standard computer vision datasets. While this limitation is also one of the strengths, leading to improved generalization capability, it requires access to auxiliary visual datasets. Lastly, our current system is still trained in a stagewise manner, with the region proposals trained on prior vision data, the attention trained from demonstration, and the policy trained from experience. An exciting direction for future work would be to enable end-to-end finetuning of the entire system, which would lift many of these limitations. Since each stage in the current method is trained with simple and scalable gradient descent methods, end-to-end training should be entirely feasible, and should improve the performance of the resulting policies on tasks that require more subtle perception mechanisms.

\bibliographystyle{IEEEtran}
\bibliography{example}

\end{document}